# If You Are Careful, So Am I! How Robot Communicative Motions Can Influence Human Approach in a Joint Task

Linda Lastrico[1,2,*], Nuno Ferreira Duarte[3], Alessandro Carfì[2], Francesco Rea[1], Fulvio Mastrogiovanni[2], Alessandra Sciutti[4], and José Santos-Victor[3]

[1] Robotics, Brain and Cognitive Science Department (RBCS), Italian Institute of Technology, Genoa, Italy
[2] Department of Informatics, Bioengineering, Robotics, and Systems Engineering (DIBRIS), University of Genoa, Genoa, Italy
[3] Institute for Systems and Robotics (ISR), Instituto Superior Técnico, Universidade de Lisboa, Lisbon, Portugal
[4] Cognitive Architecture for Collaborative Technologies Unit (CONTACT), Italian Institute of Technology, Genoa, Italy

**Abstract.** As humans, we have a remarkable capacity for reading the characteristics of objects only by observing how another person carries them. Indeed, how we perform our actions naturally embeds information on the item features. Collaborative robots can achieve the same ability by modulating the strategy used to transport objects with their end-effector. A contribution in this sense would promote spontaneous interactions by making an implicit yet effective communication channel available. This work investigates if humans correctly perceive the implicit information shared by a robotic manipulator through its movements during a dyadic collaboration task. Exploiting a generative approach, we designed robot actions to convey virtual properties of the transported objects, particularly to inform the partner if any caution is required to handle the carried item. We found that carefulness is correctly interpreted when observed in the robot movements. In the experiment, we used identical empty plastic cups; nevertheless, participants approached them differently depending on the attitude shown by the robot: humans change how they reach for the object, being more careful whenever the robot does the same. This emerging form of motor contagion is entirely spontaneous and happens even if the task does not require it.

* Corresponding author linda.lastrico@iit.it
This paper is supported by the European Commission within the Horizon 2020 research and innovation program, under grant agreement No 870142, project APRIL (multipurpose robotics for mAniPulation of defoRmable materIaLs in manufacturing processes). N. F. Duarte is supported by FCT-IST fellowship grant PD/BD/135116/2017.

# 1   Introduction

Humans routinely engage in joint actions and coordinate their movements with others, e.g. working together, playing a team sport, or merely moving objects. These tasks involve a collaborative process to coordinate attention, communication, and actions to achieve a common goal. During this process, humans observe the behavior of their partners to anticipate their actions and plan their own accordingly. Verbal communication is not the only means to express intentions. Since verbalizing every step of the interaction would be time-consuming and cognitively expensive, humans also exploit their bodies and movements to exchange information. While executing an intended action, we also implicitly communicate to others our goal, its urgency, and the required effort. This ability is referred to as non-verbal communication (i.e., non-verbal cues), and it can be expressed with our body: from turning the head or torso to a simple eye movement.

In ordinary life, humans are very proficient at monitoring different components of other people's kinematics, which they leverage to disclose hidden qualities of a handled item. For instance, studies on human non-verbal cues found that joints kinematics and dynamics of hand manipulation are crucial features to estimate the weight of a manipulated object [1,16,19] or predicting action duration [7].

Given the importance of implicit cues in human-human communication, we believe it should be taken into account also in the robotic field. A robot meant to interact with humans, able to exploit the same communication channels as the partner, would guarantee a natural and less cumbersome experience. Indeed, numerous channels of communication may be employed to convey information between robots and people (such as synthetic speech, light-based, digital display, mixed or augmented reality [5,15,13]). However, a valuable alternative that does not require any training or explicit instruction is mediated by movement, and it should be sought whenever feasible. Human non-verbal cues from eyes, head, and arm movements encode the intention driving the action; when such cues are embedded onto a robot, they similarly allow to read the robot's intention [4]. Regarding object manipulation, object affordances was popularized in robotics and linked to *(i)* the action associated with the object, *(ii)* a physical property, or *(iii)* the type of behaviour required to manipulate the object [18,9]. Specifically, works on affordance reasoning examine the object's properties [21,8,23], e.g., how to infer the water level in cups [14], although trying to directly detect such property from the object appearance, making it only possible with transparent cups and glasses.

To overcome the need to understand the properties of objects from their appearance alone, it is relevant to quantify the effect their features have on the kinematics of the action during manipulation. In our previous works, we exploited human kinematics to infer the impact of cup water content on human motion, irrespective of the cup's transparency [3,11]. Indeed, it has been shown that humans alter their behavior, adapting to the properties of the object they transport, such as weight, fragility, or content. Additionally, depending on the type of cup, these behaviors may be more predominant or less, which may indicate that the difficulty of the manipulation impacts the human motion [17]. Knowing from the mentioned studies that humans reveal some object properties through movements, in this work, we investigate if it is possible to modulate the movements adopted by a robot end-effector during the transport of an object to communicate some of its hidden properties. In a previous study [12], we assessed the communicative potential of movements on different humanoid robots, by asking participants to explicitly judge the robotic motion's carefulness after observing it in videos. This study proposes a dyadic interaction with a new robotic manipulator in a realistic collaborative context. We used Generative Adversarial Networks (GANs) to synthesize and design the robot movements to convey a particular style feature associated with object manipulation: carefulness [6]. By using a generative approach, we can consistently produce novel but meaningful robot actions. In this study, we explore (i) whether the attitude conveyed by the robot's movements is perceived as expected, i.e., if the carefulness (or its absence) is correctly expressed by our controller, and (ii) if a robot transporting objects and expressing the appropriate human-like behavior can invoke motor adaptation in the human response.

## 2   Materials and Methods

The objective of our study is to assess whether the generated robot's movements are informative of the properties of the transported object. Moreover, we evaluate if the robot behavior affects how humans perform their tasks.
We will now explain how we synthesized the required velocity profiles and controlled a Kinova Gen3 robot with 7 degrees of freedom to execute them; then, we will describe the experimental setup and design.

### 2.1   Generation of robot movements

To create robot's movements communicative about the object properties, we used Generative Adversarial Networks (GANs) and controlled the robot end-effector to follow the velocity profiles produced by such model. Our interest is in generating movements to convey whether the transported object requires caution and care to be transported (*careful* movement) or it is safe to move without any particular concern (*not careful* movement). Previous studies assessed this kind of object manipulation and showed a marked difference in the kinematics of the human hand associated with the two classes of motions [11,3]. The velocity profile is mainly affected, where actions associated with delicate objects, e.g., a cup full of water, are characterized by lower maximum velocity, prolonged deceleration phase a longer duration. This modulation is so marked that it is possible to discriminate between careful movement and not automatically [3,10].

**GAN**  To define meaningful movements associated with the properties of the carried object, we decided to modulate the velocity profile adopted by the robot

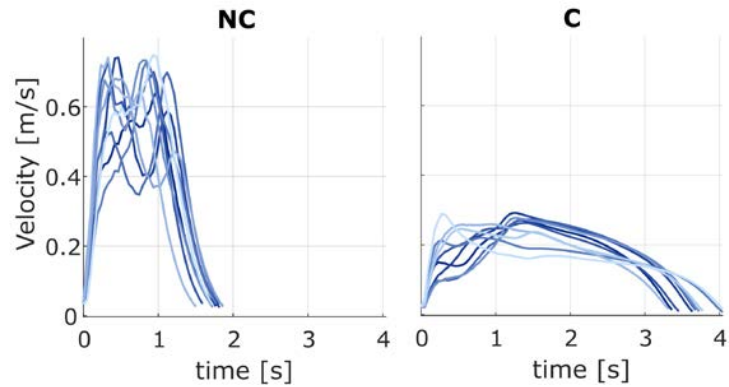

Fig. 1: Velocity profiles generate by the GANs for associated to Not Careful (NC) and Careful (C) transportation of objects. These velocity profiles were used to control the robot and provide the stimuli to the participants during the interaction experiment

end-effector. We used Generative Adversarial Networks to synthesize novel velocity profiles using an approach already tested [6]. The details on the Time-GAN model [22] and its training are described in [6]. The original data used to feed the GANs consisted of hand velocity profiles recorded with a Motion Capture System during the transport of glasses, either empty or containing water, at two possible weight levels. The trajectories followed by participants during the manipulations were designed to grant a good degree of variability. After training, each generative model can produce novel yet meaningful velocity profiles belonging to the distribution of the human data used during the training. This approach provides new and unlimited synthetic data, always falling in the desired class of motion (careful or not), avoiding a trivial copy of the human velocity profiles. Moreover, learning the velocity norm is useful in generalization terms since the same pattern of motion can be applied to multiple spatial trajectories. For this specific study, from the trained GANs, we synthesized ten velocity profiles for each of the two classes to be replicated by the Kinova robot. A representation of the generated data is available in Figure 1.

**Robot Controller** The Kinova Gen3 robot is controlled using ROS and the package kortex_ros[1]. Such package provides a velocity controller in Cartesian space, which moves the end-effector at 40 Hz in linear (m/s) and angular (rad/s) velocities. Attached to the end-effector is the Robotiq 85 two-finger gripper[2] used to grasp the cups. This work applies two high-level controllers: (i) a velocity PI controller and (ii) a velocity GAN controller. The first controller is responsible for picking the cups from the table, and the second is for transporting and handing over the cups to the participant. The former generates a constant velocity profile throughout the trials, while the latter follows one of the 20 GAN velocity profiles selected (10 careful and 10 not careful) during the experiment. For each GAN motion trajectory, the velocity profile is decomposed into the 3D Cartesian velocity coordinates by setting the current location and final location (handover point) at each time step. The handover location was fixed in advance to avoid any variability that could influence participants during the experiment. The position of the participant's wrist was tracked with a motion capture system, and the position of the robot gripper was estimated the same way (more details on the sensors used are stated in Section 2.2). The handover release moment was obtained by applying a threshold: the robot opened the gripper to release the cup whenever the distance between its end-effector and the participant's wrist was below a fixed value. This simple design was enough to grant a smooth and reactive handover required for our experiment.

---

[1] Official repository of the Kinova Gen3 ROS package: https://github.com/Kinovarobotics/ros_kortex

[2] Official website of the gripper: https://robotiq.com/products/2f85-140-adaptive-robot-gripper

## 2.2   Setup, sensors and experiment design

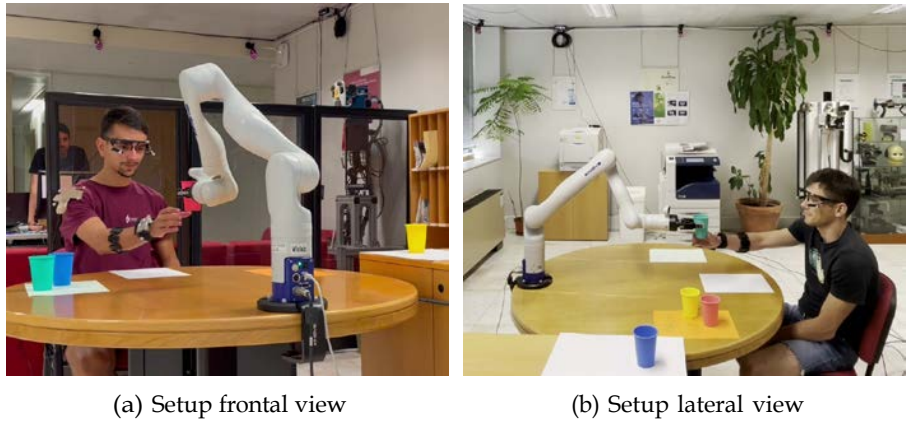

(a) Setup frontal view          (b) Setup lateral view

Fig. 2: *Setup:* when interacting with the Kinova Gen3 robot, participants seated at a table. Once grasped the cup from the robot gripper, they had to put it down on one of the three areas delimited on the table. The motion capture markers used to analyse the human kinematics are visible on the participants' right wrist

Participants were asked to sort the items handled by the Kinova Gen3 robot by positioning them on the appropriate areas marked on the table where they were seated. We designed the experiment for the participants to focus on the robot behaviour and not on the characteristics of the items. For this reason, we used identical plastic cups: in the instructions, we explained that we were simulating a bar-like scenario, where the robot and the human had to collaborate in sorting the glasses between those full to be served to the clients, and the used and empty ones, to be washed; in such context, the cups were meant to be either full of a liquid or empty: however, we explained to the participants that due to the danger of having a robot transporting water, all the cups were empty. This granted that participants could not rely on any visual cue or the actual object features to decide where to place the cup. In every trial, the Kinova robot grasped a cup from the table next to it (see Figure 2b) and transported it towards the participant, following either a careful (or not) velocity profile generated by the GANs (associated respectively, to the transport of a full or empty glass). The task for the participants was then to grasp the cup from the robot gripper and place it in the appropriate area on the table: on the "To be served" area, on the right, if they thought that the cup was actually meant to be full, or on the "To be washed" area, in case they assumed the cup was indeed empty. A third area, in the middle, was available to place the cups whose virtual content was not clear to the participant to avoid forcing them into making a decision. They were not informed about the modulation of the robot transport movements and, since the cups were all the same, they had to rely on the robot behavior to make their decision[3].

We used Optitrack[4] motion capture system, with an acquisition frequency of 120 Hz, to track the position of the participant wrist and shoulder.

Twelve healthy participants, all members of Instituto Superior Técnico, voluntarily took part in the experiment. Each evaluated 20 robot movements, where the sequence of careful and not careful modulation was randomized once and then maintained for every participant. The interactions were organized as five blocks of a sequence of four trials. At the end of each block, the experimenter put the cups on the table next to the robot. This resulted in a total of 240 movements evaluated, equally balanced between careful or not robot behavior.

## 3   Results

One of the aims of this study was to verify whether modulating the robot end-effector velocity to express carefulness can inform participants about the virtual content of the manipulated glasses. Figure 3 shows the participants' accuracy in evaluating, for each trial, if the observed transportation motion was meant to be associated with a delicate object, i.e., careful robot movement, or not. We represented with a dark bar the percentage of correct answers given by the participants, i.e., when they correctly interpreted the robot's attitude. Considering the total number of evaluated trials (240), 189 were correctly classified with no indecision, resulting in an accuracy of **78.75%**. The transparent colored bars represent the misclassified movements. For instance, when we generated a robot

---

[3] Sample video of the human-robot interaction: https://www.youtube.com/watch?v=HVahS-0tn6g

[4] Optitrack website: https://optitrack.com/cameras/flex-13/

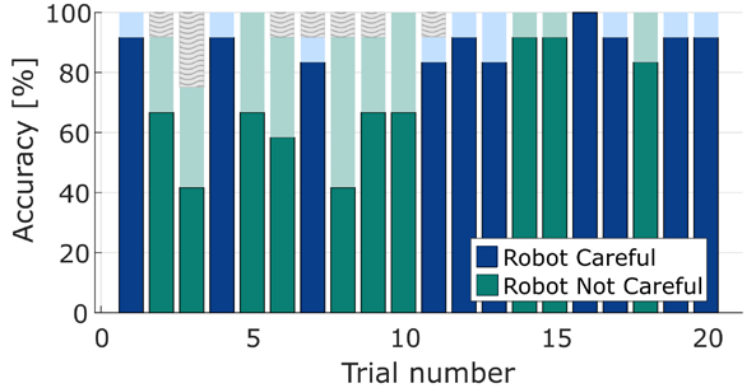

Fig. 3: *Perception of robot's movements:* Percentage of correct interpretation of the robot's transportation movements during the experiment. When the robot performed careful movements, in blue, they were correctly perceived **90%** of the times. NC motions required more trials to be consistently classified. The dark bars represent the percentage of correct classification of the movement from the participant, the transparent bars the percentage of wrong attribution; finally, the light gray bars with a wavy pattern, the percentage of "Unknown" answers in each trial

action modulated to communicate a not careful attitude, while participants associated it with the transport of a full cup. As it can be noticed, misunderstanding a not careful action for a careful one was the most frequent occurrence, especially in the first trials. In detail, **90%** of the careful robot movements were perceived as such, whereas **75%** of the not careful ones were correctly interpreted. Finally, the grey bars with a wavy pattern represent those trials where participants preferred not to make a choice and placed the cup on the neutral area on the table. Also, these occurrences, which happened in 9 trials out of 240, decrease as the experiment progresses.

Another aspect we were interested in investigating is if the two attitudes shown by the robot had any effect on how participants performed their tasks. An exploratory inspection of the hand velocity data encouraged us to deepen this intuition: Figure 4 reports an example of the velocity adopted by one participant when reaching for the cup in the robot gripper. There is a noticeable modulation in the participant's movements that correlates with the attitude shown by the robot. When the robot handled the cup with a careful attitude, also the participant reached for it with slower and prolonged action compared to the not careful situation. To assess this modulation quantitatively in human actions, we considered the duration and median velocity of the movements as relevant features. To perform statistical analyses on the acquired data, we used

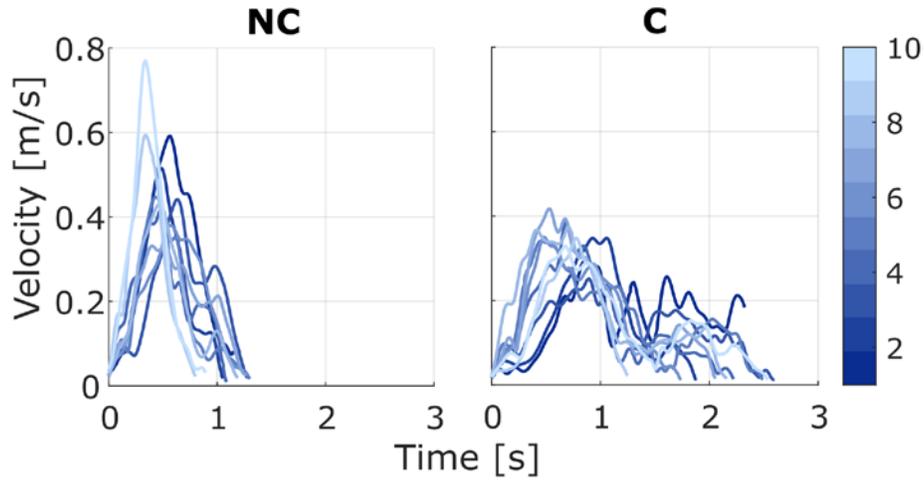

Fig. 4: *Velocity reaching movement:* profiles adopted by one participant when reaching for the cup in the robot gripper. It is noticeable a modulation of both the duration and the maximum values depending on the style of the movement adopted by the robot: not careful (NC) or careful (C). The colormap is associated to the trial numbers, in order

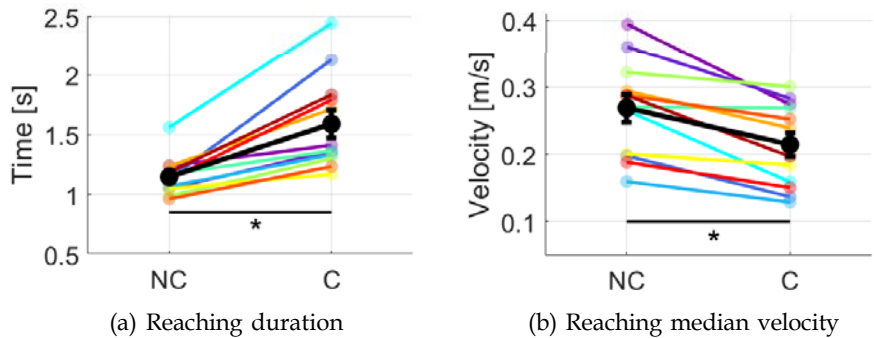

(a) Reaching duration          (b) Reaching median velocity

Fig. 5: *Reaching movement:* in (5a) mean duration of the participants reaching movements towards the robot's gripper. When the robot performs a Careful (C) transportation movement, participants are significantly slower in reaching for the cup. Also the median velocity adopted in the reaching movements (5b) is modulated by how the robot moved in the transport action. The mean values for each participant are represented in a different color. The thick black lines represent the mean over the twelve participants, with the standard error. The star indicates a significant difference with $p < 0.001$

Jamovi software[5], in particular the GAMLj module[6] for mixed models. Figure 5a shows the mean durations of the participants reaching movements toward the robot gripper. We ran a mixed model assuming the duration of the participants' reaching movements as the dependent variable, the carefulness in the robot movement as a factor, and the subjects as cluster variables. The effect of condition resulted significant ($C - NC$, $estimate = 0.443$, $SE = 0.055$, $t = 8.00$, $p < 0.001$), indicating that when the robot end-effector was following a careful velocity profile, the subsequent human reaching action was longer, with an extended duration estimate of 0.443 seconds. A second mixed model was used to evaluate the median velocity adopted by the participants when reaching the robot gripper (see Figure 5b), using this time the median velocity as dependent variable: when the robot was careful, participants significantly diminished their median velocity, with an estimated reduction in speed of $0.055 m/s$ ($C - NC$, $estimate = 0.055$, $SE = 0.012$, $t = -4.60$, $p < 0.001$). These findings prove that the modulation of the robot movements affected how participants moved to reach the cup and take it from the robot gripper. This happened even if there was no reason to adapt to the object properties since we consider a reaching movement without any object directly involved; moreover, all the cups had exactly the same characteristics. We also verified, for both the duration and the median velocity of the reaching movements, if there was an interaction with the participants' accuracy in evaluating the robot's behavior in every trial. We used the accuracy in their classification as an additional factor in the mixed model, but we found no interaction with how they performed the reaching duration or velocity. The modulation in response to the robot attitude also occurred when participants did not recognize it explicitly.

## 4  Discussion

In our study, we exploited a generative approach to produce robot movements that could implicitly communicate if a handled object required or not carefulness to be transported. To avoid influencing the choice, all the items transported by the Kinova robot were identical (empty plastic cups). The participants had to decide if they were supposed to be virtually full or empty, without any particular hint or instruction on how to proceed. Firstly, we assessed (i) whether our controller can express caution in the gestures or its absence. According to the results shown in Figure 3, we notice that the *careful* robot actions have been perceived as such since the first trials of the experiment. Regarding the *not careful* actions, there is a learning curve in how they were perceived during the experiment. In the first trials, they were sometimes mistaken for actions associated with transporting a full cup. As the experiment progressed, the difference between the two modulations became more evident, with an accuracy in the participants' choices above 80%. Reflecting on the original dataset of human movements used to train the GANs, associated with the transport of full and empty cups [11],

---

we can observe that a *not careful* attitude is standard in our actions. Indeed, when no particular circumstances are forcing us, for instance when picking and placing an ordinary object, we tend to move in a "neutral" way, and we can shortly describe our gesture as not careful. On the contrary, a strong kinematics modulation appears when we are paying attention to not spill the content of a glass [3,10]. This careful kinematics shaping is what we truly modeled in the communicative robot's movements, and it is rewarding that careful movements were perceived correctly from the beginning.

This study also allows us to evaluate (ii), the effect that the implicit modulation of the robot actions has on the interaction. Even though participants knew from the beginning that the plastic cups were all the same and all empty, there was a modulation in how they approached the robot gripper. We gave an overview of this phenomenon in Figure 4 and a quantitative assessment in Figure 5. If the robot manifested a careful attitude, adopting a lower magnitude in the velocity profile and a longer duration of the movement (see Figure 1 for reference), also the reaching movement of the humans was significantly slower. Interestingly, this also happened when participants had trouble explicitly recognizing the motion style and classifying the cup: the contagion in how they performed the reaching task was still present. This result emphasizes how important it is to modulate the actions of robots appropriately, with a view to collaborative interaction. Indeed, we proved a motor contagion from the robot to the human, even if there was for the participants no need directly associated with the task to adapt their motor strategies. We observed natural coordination emerging from such a simple task, where the pace of the human spontaneously adapted to the robot one, mimicking, even unconsciously, the attitude observed. Human-robot motor contagion on velocity was already observed, as far as the robot velocity profile is biologically plausible [2,20]. In our approach, the reasonableness of the velocity profiles was granted using a generative network trained on human examples. The findings in our study extend the existing evidence of motor contagion in Human-Robot Interaction, proving that robotics arms can also leverage it to convey appropriate ways of handling fragile objects.

## 5    Conclusion

In this study, we showed how a generative approach could be used to generate meaningful and communicative robot actions that a human partner can successfully interpret to infer some properties of the involved objects. This modulation on the robot side also led to a motor contagion in how the human performed its actions and synchronized with the pace of the robot; through motion alone, it was possible to open a channel of communication between the two agents, with measurable effects on the interaction.

Finally, it should be noted that we obtained these results by modulating the movements of a 7 degrees of freedom robotic manipulator, not a humanoid robot. Nevertheless, even though its kinematics was far from the one of a human arm, it was possible to achieve the desired communication intent by simply modulating the end-effector control. This proves the power of the proposed approach and its potential scalability in other contexts and with other robots, also industrial ones, where implicit communication through motion could improve the efficiency and safety of a joint collaborative task.

In future works, we plan to exploit the same controller and have the robot actually manipulate full and empty cups to assess how the movement's modulation affects trust, perceived competency, and efficiency in a dyadic interaction, while facing a challenging task.